\pdfoutput=1
\documentclass[11pt]{article}

\usepackage[final]{acl}

\usepackage{times}
\usepackage{latexsym}
\usepackage{booktabs}
\usepackage{graphicx}
\usepackage{array}
\usepackage{amsmath}
\usepackage{graphicx}
\usepackage{caption}
\usepackage{subcaption}
\usepackage{comment}
\usepackage{arydshln}
\usepackage{upgreek}
\usepackage{multirow}
\usepackage{mathtools}
\usepackage[utf8]{inputenc} % allow utf-8 input
\usepackage[T1]{fontenc}    % use 8-bit T1 fonts
\usepackage{url}            % simple URL typesetting
\usepackage{booktabs}       % professional-quality tables
\usepackage{amsfonts}       % blackboard math symbols
\usepackage{nicefrac}       % compact symbols for 1/2, etc.
\usepackage{microtype}      % microtypography
\usepackage{xcolor}         % colors

\usepackage[T1]{fontenc}
\usepackage[utf8]{inputenc}

\usepackage{microtype}

\usepackage{inconsolata}

\usepackage{graphicx}
\newcommand{\intel}{$^{\heartsuit}$}
\newcommand{\oracle}{$^{\clubsuit}$}
\newcommand{\tw}{$^{\diamondsuit}$}

\title{Pruning the Paradox: How CLIP’s Most Informative Heads Enhance Performance While Amplifying Bias}

\author{
  Avinash Madasu\intel \qquad
  Vasudev Lal\oracle\thanks{Work partially completed while at Intel Labs.} \qquad
  Phillip Howard\tw\footnotemark[1] \\
  \intel Intel Labs \qquad
  \oracle Oracle \qquad
  \tw Thoughtworks \\
  \qquad {avinash.madasu@intel.com} \qquad phillip.howard@thoughtworks.com
}

\begin{document}
\maketitle
\begin{abstract}
CLIP is one of the most popular foundation models and is heavily used for many vision-language tasks, yet little is known about its inner workings. As CLIP is increasingly deployed in real-world applications, it is becoming even more critical to understand its limitations and embedded social biases to mitigate potentially harmful downstream consequences. However, the question of what internal mechanisms drive both the impressive capabilities as well as problematic shortcomings of CLIP has largely remained unanswered. To bridge this gap, we study the conceptual consistency of text descriptions for attention heads in CLIP-like models. Specifically, we propose Concept Consistency Score (CCS), a novel interpretability metric that measures how consistently individual attention heads in CLIP models align with specific concepts. Our soft-pruning experiments reveal that high CCS heads are critical for preserving model performance, as pruning them leads to a significantly larger performance drop than pruning random or low CCS heads. Notably, we find that high CCS heads capture essential concepts and play a key role in out-of-domain detection, concept-specific reasoning, and video-language understanding. Moreover, we prove that high CCS heads learn spurious correlations which amplify social biases. These results position CCS as a powerful interpretability metric exposing the paradox of performance and social biases in CLIP models.
\end{abstract}

\section{Introduction}

Large-scale vision-language (VL) models such as CLIP ~\cite{radford2021learning} have significantly advanced state-of-the-art performance in vision tasks in recent years. Consequently, CLIP has been extensively used as a foundational model for downstream tasks such as video retrieval, image generation, and segmentation ~\cite{luo2022clip4clip, liu2024improved, brooks2023instructpix2pix, esser2024scaling, kirillov2023segment}. This has enabled the construction of compositional models combining CLIP with other foundation models, thereby increasing the functionality of CLIP while also adding complexity to the overall model structure. However, as these models gain prominence in real-world applications, their embedded social  biases ~\cite{howard2024socialcounterfactuals, hall2023visogender, seth2023dear} have emerged as a critical concern with potentially harmful consequences. Despite the growing body of work documenting these biases, a fundamental question remains: \textit{what mechanisms within these models' architectures drive both their impressive capabilities and problematic shortcomings?}

Recent interpretability advances ~\cite{gandelsmaninterpreting} have made initial progress by decomposing CLIP's image representations into contributions from individual attention heads, identifying text sequences that characterize different heads' semantic roles. However, this approach provides only a partial view into CLIP's inner workings, leaving a critical missing piece: systematic understanding of the visual concepts encoded at the attention head level—and how these concepts underpin both the model’s strengths and its social failures. 

Our work addresses this critical gap through a novel interpretability framework we call ``conceptual consistency''. This framework systematically analyzes which visual concepts are learned by individual attention heads and how consistently these concepts are processed throughout the model's architecture. First, we identify interpretable structures within the individual heads of the last four layers of the model using a set of text descriptions. To accomplish this, we employ the \textsc{TextSpan} algorithm ~\cite{gandelsmaninterpreting}, which helps us find the most appropriate text descriptions for each head. After identifying these text descriptions, we assign labels to each head representing the common property shared by the descriptions. This labeling process is carried out using in-context learning with ChatGPT. We begin by manually labeling five pairs of text descriptions and their corresponding concept labels, which serve as examples. These examples are then used to prompt ChatGPT to assign labels for the remaining heads.

Leveraging the resulting text descriptions and concept labels of attention heads, we introduce the Concept Consistency Score (CCS), a new interpretability metric that quantifies how strongly individual attention heads in CLIP models align with specific concepts. Using GPT-4o, Gemini and Claude as automatic judges, we compute CCS for each head and classify them into high, moderate, and low categories based on defined thresholds. A key contribution of our work is our targeted soft-pruning experiments which show that heads with high CCS are essential for maintaining model performance; pruning these heads causes a significantly larger performance drop compared to pruning any other heads. We also show that high CCS heads are not only crucial for general vision-language tasks but are especially important for out-of-domain detection and targetted concept-specific reasoning. Additionally, our experiments in video retrieval highlight that high CCS heads are equally vital for temporal and cross-modal understanding. Moreover, we demonstrate that high CCS heads often encode spurious correlations, contributing to social biases in CLIP models. Selective pruning of these heads can reduce such biases without the need for fine-tuning. Together, these results expose a fundamental paradox: while high-CCS heads are indispensable for strong model performance, they are simultaneously key contributors to undesirable biases.

\begin{figure*}[h!]
    \centering
    \includegraphics[width=0.97\linewidth]{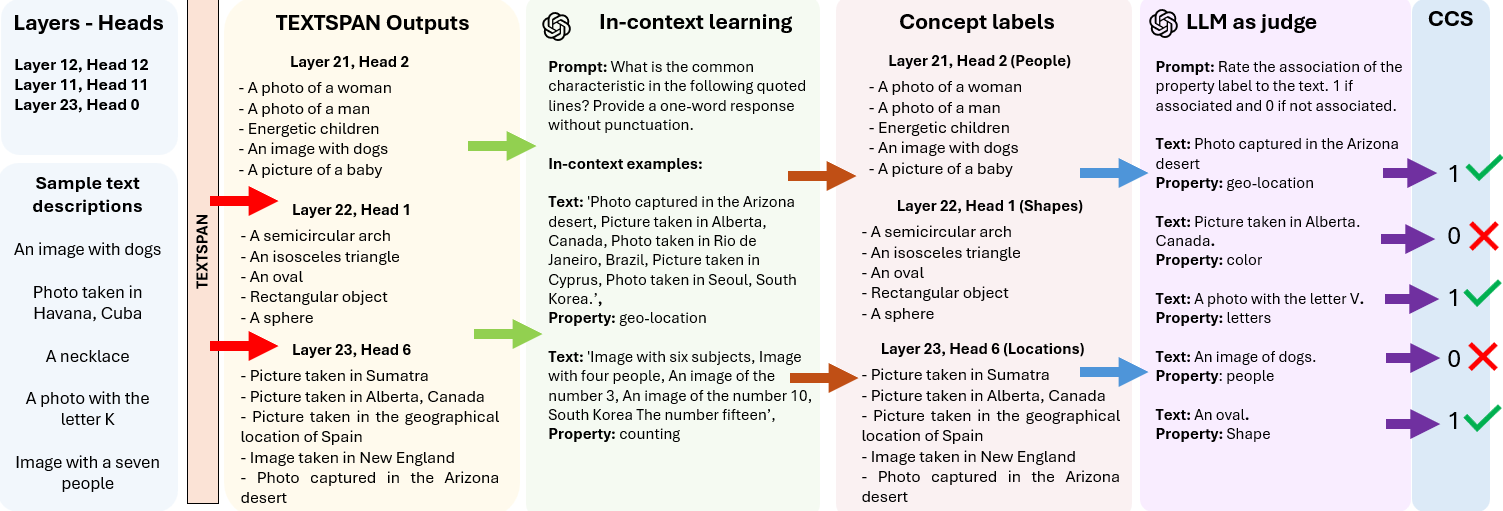}
    \caption{\textbf{Illustration of our approach to computing Concept Consistency Score for each attention head.}}
    \label{fig:llm-as-judge}
\end{figure*}

\begin{table*}[h!]
\small
\centering
\renewcommand{\arraystretch}{1.0}
\begin{tabular}{p{4.6cm}|p{4.6cm}|p{4.6cm}}
\hline
\textbf{High CCS $(CCS = 5)$} & \textbf{Moderate CCS} $(CCS = 3)$ & \textbf{Low CCS $(CCS \leq 1)$} \\
\hline
\hline
\textbf{L23.H11} (``People'') & \textbf{L23.H0 (``Material'')} & \textbf{L21.H6 (``Professions'')} \\
\hline
Playful siblings & Intrica wood carvingte & Photo taken in the Italian pizzerias \\
A photo of a young person & Nighttime illumination & thrilling motorsport race \\
Image with three people & Image with woven fabric design & Urban street fashion \\
A photo of a woman & Image with shattered glass reflections & An image of a Animal Trainer \\
A photo of a man & A photo of food & A leg \\

\hline
\textbf{L22.H10} (``Animals'') & \textbf{L11.H0 (``Locations'')} & \textbf{L10.H6 (``Body parts'')} \\
\hline
Image showing prairie grouse & Photo taken in Monument Valley
 & A leg \\
Image with a donkey & Majestic animal & colorful procession \\
Image with a penguin & An image of Andorra & Contemplative monochrome portrait \\
Image with leopard print patterns & An image of Fiji & Graceful wings in motion \\
detailed reptile close-up & Image showing prairie grouse & Inviting reading nook \\

\hline
\textbf{L23.H5} (``Nature'') & \textbf{L11.H11 (``Letters'')} & \textbf{L9.H2 (``Textures'')} \\
\hline
Intertwined tree branches & A photo with the letter J & Photo of a furry animal \\
Flowing water bodies & A photo with the letter K & Closeup of textured synthetic fabric \\
A meadow & A swirling eddy & Eclectic street scenes \\
A smoky plume & A photo with the letter C & Serene beach sunset \\
Blossoming springtime blooms & awe-inspiring sky & Minimalist white backdrop \\

\hline
\end{tabular}
\caption{\textbf{Examples of high, moderate and low CCS heads.}}
\label{tab:ccs_table}
\end{table*}

\begin{table}[]
\small
    \centering
    \begin{tabular}{cccc}
    \toprule
    \textbf{Model} & \textbf{High} & \textbf{Medium} & \textbf{Low} \\
    \midrule
      ViT-B-32-OpenAI  & 8 & 13 & 4 \\
ViT-B-32-datacomp & 11 & 9 & 8 \\
\midrule
ViT-B-16-OpenAI & 10 & 14 & 4 \\
ViT-B-16-LAION & 13 & 12 & 6 \\
\midrule
ViT-L-14-OpenAI & 16 & 13 & 11 \\
ViT-L-14-LAION & 21 & 14 & 3 \\
\bottomrule
    \end{tabular}
    \caption{\textbf{Count of high, medium and low CCS heads in CLIP models.}}
    \label{tab:head-counts}
\end{table}

\begin{table*}[htbp]
\centering
\resizebox{\textwidth}{!}{%
\begin{tabular}{|l|cccccc|cccccc|}
\hline
\multirow{2}{*}{\textbf{Model}} &
\multicolumn{6}{c|}{\textbf{CIFAR-10}} &
\multicolumn{6}{c|}{\textbf{CIFAR-100}} \\
& \textbf{Original} & \textbf{K=5} & \textbf{K=7} & \textbf{K=9} & \textbf{K=11} & \textbf{K=13}
& \textbf{Original} & \textbf{K=5} & \textbf{K=7} & \textbf{K=9} & \textbf{K=11} & \textbf{K=13} \\
\hline
ViT-B-32-OpenAI     & 75.68 & 71.31 & 71.86 & 73.66 & 74.53 & 70.54  & 65.08 & 56.07 & 62.21 & 60.17 & 56.42 & 59.05 \\
ViT-B-32-datacomp   & 72.07 & 70.50 & 70.59 & 69.61 & 69.93 & 70.20  & 54.95 & 53.14 & 53.59 & 53.73 & 53.81 & 53.91 \\
ViT-B-16-OpenAI     & 78.10 & 63.93 & 71.83 & 76.08 & 64.67 & 76.74  & 68.22 & 51.70 & 52.45 & 61.90 & 62.59 & 55.54 \\
ViT-B-16-LAION      & 82.82 & 78.91 & 81.25 & 80.25 & 78.03 & 78.56  & 76.92 & 65.55 & 69.94 & 72.46 & 67.59 & 69.49 \\
ViT-L-14-OpenAI     & 86.94 & 86.29 & 85.44 & 85.89 & 85.49 & 85.58  & 78.28 & 75.66 & 74.85 & 74.70 & 76.73 & 77.19 \\
ViT-L-14-LAION      & 88.29 & 86.48 & 85.99 & 86.34 & 87.28 & 85.99  & 83.37 & 80.07 & 80.75 & 82.03 & 82.14 & 80.11 \\
\hline
\end{tabular}
}
\caption{\textbf{Accuracy (\%) on CIFAR-10 and CIFAR-100 datasets across different values of \textsc{TEXTSPAN} ($K$).}}
\label{tab:text-spans}
\end{table*}

\section{Related Work}
\label{sec:formatting}

Early research on interpretability primarily concentrated on convolutional neural networks (CNNs) due to their intricate and opaque decision-making processes \citep{zeiler2014visualizing,selvaraju2017grad,simonyan2014deep,fong2017interpretable,hendricks2016generating}. 
% Zeiler et al. introduced deconvolutional networks, enabling the visualization of CNN activations to pinpoint which parts of an image influenced the model's predictions \citep{zeiler2014visualizing}. Subsequently, Selvaraju et al. proposed Grad-CAM, a technique that employs gradients to create class activation maps highlighting significant regions in an image \citep{selvaraju2017grad}. In addition saliency maps were proposed which utilize gradients to reveal important features within images \citep{simonyan2014deep}. Perturbation-based methods were introduced for visualizing and understanding feature importance \citep{fong2017interpretable}, while Hendricks et al. expanded these ideas by producing localized explanations for classifiers, focusing on key image regions \citep{hendricks2016generating}. These early contributions established a strong foundation for interpretability techniques in CNNs. 
More recently, the interpretability of Vision Transformers (ViT) has garnered significant attention as these models, unlike CNNs, rely on self-attention mechanisms rather than convolutions. Researchers have focused on task-specific analyses in areas such as image classification, captioning, and object detection to understand how ViTs process and interpret visual information \cite{dong2022towards, elguendouze2023explainability, mannix2024scalable, xue2022protopformer, cornia2022explaining, dravid2023rosetta}. One of the key metrics used to measure interpretability in ViTs is the attention mechanism itself, which provides insights into how the model distributes focus across different parts of an image when making decisions \cite{cordonnier2019relationship, chefer2021transformer}. This has led to the development of techniques that leverage attention maps to explain ViT predictions. Early work on multimodal interpretability, which involves models that handle both visual and textual inputs, probed tasks such as how different modalities influence model performance \cite{cao2020behind, madasu2023multimodal} and how visual semantics are represented within the model \cite{hendricks2021probing, lindstrom2021probing}. Aflalo et al. \citep{aflalo2022vl} explored interpretability methods for vision-language transformers, examining how these models combine visual and textual information to make joint decisions. Similarly, Stan et al. \citep{stan2024lvlm} proposed new approaches for interpreting vision-language models, focusing on the interactions between modalities and how these influence model predictions. Our work builds upon and leverages the methods introduced by \citet{gandelsmaninterpreting,gandelsman2024neurons} to interpret attention heads, neurons, and layers in vision-language models, providing deeper insights into their decision-making processes. Several works ~\cite{shu2023clipood, mayilvahanansearch, moeller2024explaining} explored the out-of-domain generalization of CLIP models on several benchmarks. In contrast to these works, our work provides a new interpretability metric to understand the decision making processes of attention heads in CLIP models.

\begin{table}[t]
\footnotesize
    \centering
    {
    \begin{tabular}{cccc}
    \toprule
     \textbf{Model} & \textbf{Kappa} & \textbf{SC ($\rho$)} & \textbf{Kendall ($\uptau$)} \\
    \midrule
    ViT-B-32-OpenAI  & 0.821 & 0.737 & 0.781  \\
ViT-B-16-LAION &  0.813 & 0.773 & 0.737   \\
ViT-L-14-OpenAI & 0.827 & 0.751 & 0.758   \\
    \bottomrule
    \end{tabular}
    }
    \caption{\textbf{Results between human judgment and LLM judgment on CCS labelling}. SC denotes Spearman's correlation.} 
    \label{tab:human-llm-eval-study}
\end{table}

\section{Quantifying interpretability in CLIP models}
\subsection{Preliminaries}
In this section, we describe our methodology, starting with the \textsc{TextSpan} ~\cite{gandelsmaninterpreting} algorithm and its extension across all attention heads in multiple CLIP models using in-context learning. Specifically, we deal with attention heads in the image encoder. \textsc{TextSpan} associates each attention head with relevant text descriptions by analyzing the variance in projections between head outputs and candidate text representations. Through iterative projections, it identifies distinct components aligned with different semantic aspects. While effective at linking heads to descriptive text spans, \textsc{TextSpan} does not assign explicit concept labels. We inherited the set of 3498 text spans from Gandelsman et al. ~\cite{gandelsmaninterpreting}, which were generated by prompting ChatGPT-3.5 to produce general image descriptions using the prompts shown below. After obtaining an initial set, ChatGPT was manually prompted to generate more examples of specific patterns we found (e.g., image with a lunar eclipse, Oceanic coral reef). In the next section, we detail our method for labeling the concepts learned by individual CLIP heads.

\subsection{Concept Consistency Score (CCS)}
We introduce the Concept Consistency Score (CCS) as a systematic metric for analyzing the concepts (properties) learned by transformer layers and attention heads in CLIP-like models. This score quantifies the alignment between the textual representations produced by a given head and an assigned concept label. Existing interpretability methods ~\cite{chefer2021transformer, abnar2020quantifying, barkan2021grad} are largely descriptive and do not identify which internal components (e.g., attention heads) are reliably responsible for model behavior. This lack of discriminative insight limits their utility in scenarios where intervention or control is required. CCS addresses this by measuring concept consistency across heads, enabling us to isolate specialized heads that consistently encode particular concepts. This makes interpretability actionable by supporting tasks like pruning, debugging, or understanding failure modes with certainty and control. Figure ~\ref{fig:llm-as-judge} illustrates our approach, with the following sections detailing each step in computing CCS.
\subsubsection{Extracting Text Representations}
From each layer and attention head of the CLIP model, we obtain a set of five textual outputs, denoted as \(\{T_1, T_2, T_3, T_4, T_5\}\), referred to as \textsc{TextSpan}s. These outputs serve as a textual approximation of the concepts encoded by the head.
\subsubsection{Assigning Concept Labels}
Using in-context learning with ChatGPT, we analyze the set of five \textsc{TextSpan} outputs and infer a concept label $C_{h}$ that best represents the dominant concept captured by the attention head 
$h$. This ensures that the label is data-driven and reflects the most salient pattern learned by the head.
\subsubsection{Evaluating Concept Consistency}
To assess the consistency of a head with respect to its assigned concept label, we employ three state-of-the-art foundational models, GPT-4o, Gemini 1.5 pro and Claude Sonnet as external evaluators. For each \textsc{TextSpan} $T_{i}$ associated with head $h$, GPT-4o determines whether it aligns with the assigned concept $C_{h}$. The Concept Consistency Score (CCS) for head $h$ is then computed as:

\[
\text{CCS}(h) = \sum_{i=1}^{5} \mathbb{1} \left[ T_i \text{ aligns with } C_h \right]
\]
where \(\mathbb{1}[\cdot]\) is an indicator function that returns 1 if \(T_i\) to be consistent with \(C_h\), and 0 otherwise. To ensure a high standard of reliability, we define consistency strictly—only if all three LLM judges independently rate \(T_i\) as consistent with \(C_h\). This requirement for unanimous agreement minimizes the influence of individual model biases or variability in judgment ~\cite{liu2025is}, thereby enhancing the robustness and trustworthiness of the overall concept consistency score.

We define $CCS@K$ as the fraction of attention heads in a CLIP model that have a Concept Consistency Score (CCS) of 
$K$. This metric provides a global measure of how many heads strongly encode interpretable concepts. A higher $CCS@K$ value indicates that a greater proportion of heads exhibit strong alignment with a single semantic property. Mathematically, $CCS@K$ is defined as:
\[
CCS@K = \frac{1}{H} \sum_{h=1}^{H} \mathbb{1} \left[ \text{CCS}(h) = K \right]
\]

where \(H\) is the total number of attention heads in the model,  
 \(\text{CCS}(h)\) is the Concept Consistency Score of head \(h\),  
\(\mathbb{1}[\cdot]\) is an indicator function that returns 1 if \(\text{CCS}(h) = K\), and 0 otherwise. This metric helps assess the overall interpretability of the model by quantifying the proportion of heads that consistently capture well-defined concepts. Table ~\ref{tab:ccs_table} shows the examples of heads with different CCS scores.

Next, we categorize each attention head based on its Concept Consistency Score (CCS) into three levels: high, moderate, and low. A head is considered to have a high CCS if all of its associated text descriptions align with the labeled concept, indicating that the head is highly specialized and likely encodes features relevant to that concept. Moderate CCS heads exhibit partial alignment, with three out of five text descriptions matching the concept label, suggesting that they capture the concept to some extent but not exclusively. In contrast, low CCS heads have zero or only one matching description, implying minimal relevance and indicating that these heads are largely unrelated to the given concept. This categorization provides insight into the degree of concept selectivity exhibited by individual attention heads. Table ~\ref{tab:ccs_table} shows examples of different types of CCS heads and Table ~\ref{tab:head-counts} shows the count of high, moderate and low CCS heads in all CLIP models.
\subsection{Evaluating LLM Judgment Alignment with Human Annotations}
In the previous section, we introduced the Concept Consistency Score (CCS), computed using three LLM judges as an external evaluator. This raises an important question: \textit{Are LLM evaluations reliable and aligned with human assessments?} To investigate this, we conducted a human evaluation study comparing LLM-generated judgments with human annotations. We selected 100 \textsc{TextSpan} descriptions from three different models, along with their assigned concept labels, and asked one of the authors to manually assess the semantic alignment between each span and its corresponding label.

Table~\ref{tab:human-llm-eval-study} reports the agreement metrics between human and LLM evaluations, including Cohen’s Kappa, Spearman’s $\rho$, and Kendall’s $\uptau$. The Kappa values exceed 0.8, indicating extremely substantial agreement, while the correlation scores consistently surpass 0.7, confirming strong alignment. These results validate the use of LLMs as reliable evaluators in concept consistency analysis. The high agreement with human judgments suggests that LLMs can effectively assess semantic coherence, offering a scalable alternative to manual annotation. In the next section, we introduce the tasks and datasets used in our experiments.

\subsection{Sensitivity of CCS to the number of \textsc{TextSpan} descriptions}
Next, we analyze the robustness of CCS computation to number of \textsc{TextSpan} ($K$) descriptions. To directly address this issue, we conduct experiments evaluating CCS performance across different numbers of TextSpan descriptions on CIFAR-10 and CIFAR-100 datasets. The results are depicted in the table ~\ref{tab:text-spans}. From the results, it is evident that performance remains remarkably stable across all $K$ values, with variations typically within 1-3\% accuracy points across multiple CLIP models. Importantly, the relative performance ordering between different model configurations remains consistent regardless of $K$, indicating that our concept alignment assessment captures meaningful patterns. Performance does not consistently improve with higher $K$ values, suggesting that five \textsc{TextSpan} descriptions capture sufficient conceptual diversity without unnecessary computational overhead. These findings directly validate that CCS is robust to the number of \textsc{TextSpan} descriptions and not overly sensitive to this hyperparameter choice.

\begin{table*}[h]
    \scriptsize
    \centering
    \resizebox{1\textwidth}{!} {
    \begin{tabular}{l| ccc| ccc | ccc}
        \toprule
        \multirow{2}{*}{\textbf{Model}} & \multicolumn{3}{c}{\textbf{CIFAR-10}} & \multicolumn{3}{c}{\textbf{CIFAR-100}} & \multicolumn{3}{c}{\textbf{FOOD-101}} \\
        \cmidrule(lr){2-4} \cmidrule(lr){5-7} \cmidrule(lr){8-10}
        & \textbf{Original} & \textbf{High CCS} & \textbf{Low CCS} & \textbf{Original} & \textbf{High CCS} & \textbf{Low CCS} & \textbf{Original} & \textbf{High CCS} & \textbf{Low CCS} \\
        \midrule
        ViT-B-32-OpenAI & 75.68 & 71.31 & 73.61 & 65.08 & 56.07 & 62.39 & 84.01 &	73.42	& 82.12 \\
       ViT-B-32-datacomp & 72.07 & 70.50 & 70.43 & 54.95 & 53.14 & 53.72 & 41.66 &	38.13 & 40.77 \\
        \midrule
        ViT-B-16-OpenAI & 78.10 & 63.93 & 76.44 & 68.22 & 51.70 & 65.38 & 88.73 & 76.35 & 87.36\\
      ViT-B-16-LAION  & 82.82 & 78.91 & 75.38 & 76.92 & 65.55 & 72.51 & 86.63 & 67.54 & 81.4\\
        \midrule
        ViT-L-14-OpenAI & 86.94 & 86.29 & 85.97 & 78.28 & 75.66 & 77.55 & 93.07 & 90.75 & 92.79\\
      ViT-L-14-LAION  & 88.29 & 86.48 & 88.19 & 83.37 & 80.07 & 83.25 & 91.02 & 86.45 & 90.35 \\
        \bottomrule
    \end{tabular}
    }
    \caption{\textbf{Accuracy comparison of various CLIP models on CIFAR-10, CIFAR-100 and FOOD-101 datasets. The values represent original accuracy, performance after pruning high-CCS heads, and performance after pruning low-CCS heads.}}
    \label{tab:cifar-food}
\end{table*}

\subsection{Experimental Setting}
\subsubsection{Tasks}
For our experiments, we use a wide variety of datasets focusing on the tasks of image classification, out-of-domain classification, video retrieval, and bias measurement. Below, we mention datasets for each of the tasks. \\
\textbf{Image classification:} CIFAR-10 ~\cite{krizhevsky2009learning}, CIFAR-100 ~\cite{krizhevsky2009learning}, Food-101 ~\cite{bossard2014food}, Country-211 ~\cite{radford2021learning} and Oxford-pets ~\cite{parkhi2012cats}. \\
\textbf{Out-of-domain classification:} Imagenet-A ~\cite{hendrycks2021natural} and Imagenet-R ~\cite{hendrycks2021many}. \\
\textbf{Video retrieval:} MSRVTT ~\cite{xu2016msr}, MSVD ~\cite{chen2011collecting}, DiDeMo ~\cite{anne2017localizing}. \\
\textbf{Bias:} FairFace ~\cite{karkkainen2021fairface}, SocialCounterFactuals ~\cite{howard2024socialcounterfactuals}.
\subsubsection{Models}
For experiments we use the following six foundational image-text models: ViT-B-32, ViT-B-16 and ViT-L-14 pretrained from OpenAI-400M ~\cite{radford2021learning} and LAION2B ~\cite{schuhmann2022laion}. 
Next, we discuss in detail the results from the experiments.

\begin{figure*}[ht]
    \centering
    % First Figure
    \begin{subfigure}[b]{0.33\textwidth}
        \centering
        \includegraphics[width=\textwidth]{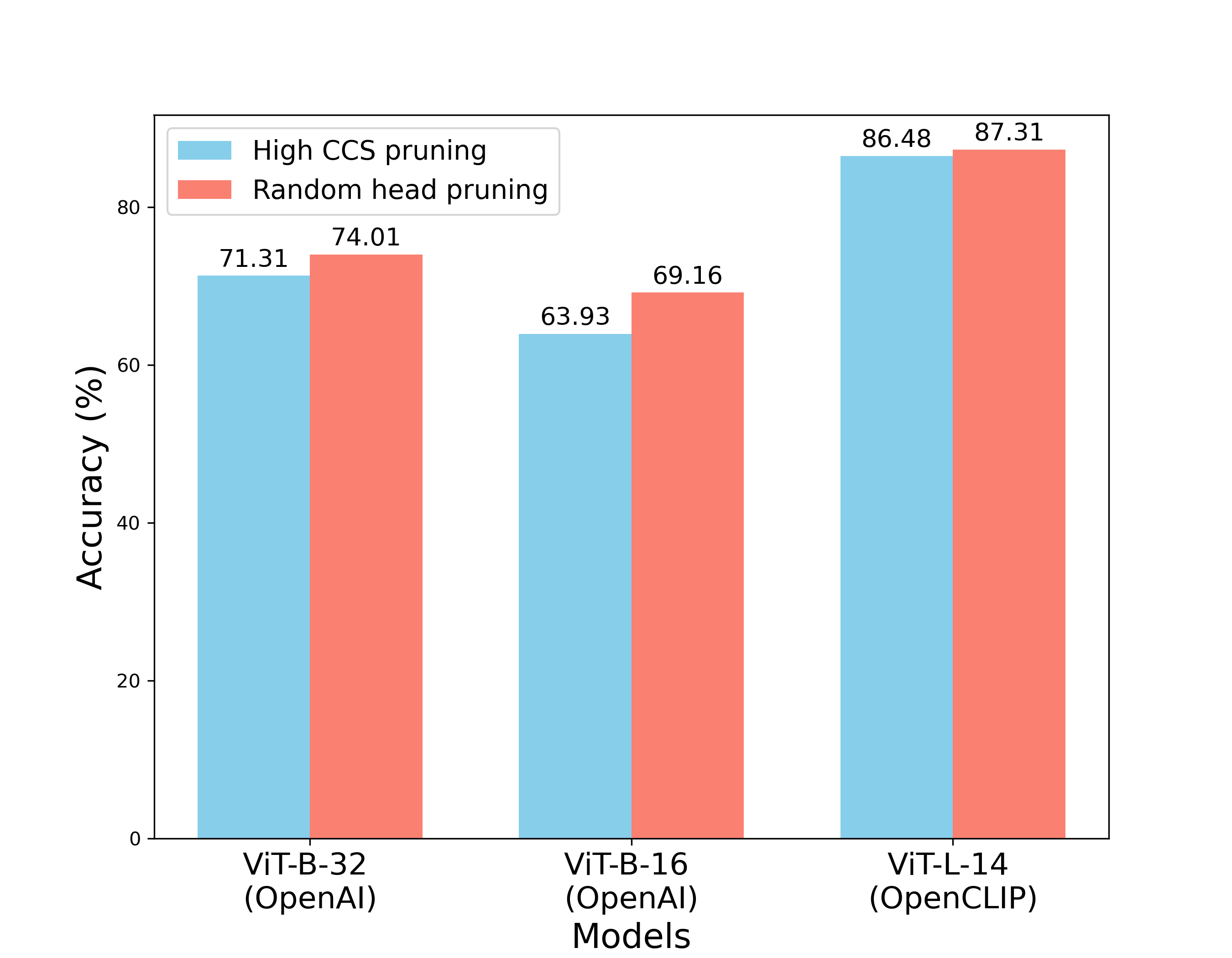}
        \caption{CIFAR-10}
        \label{fig:cifar10}
    \end{subfigure}
    \hfill
    % Second Figure
    \begin{subfigure}[b]{0.33\textwidth}
        \centering
        \includegraphics[width=\textwidth]{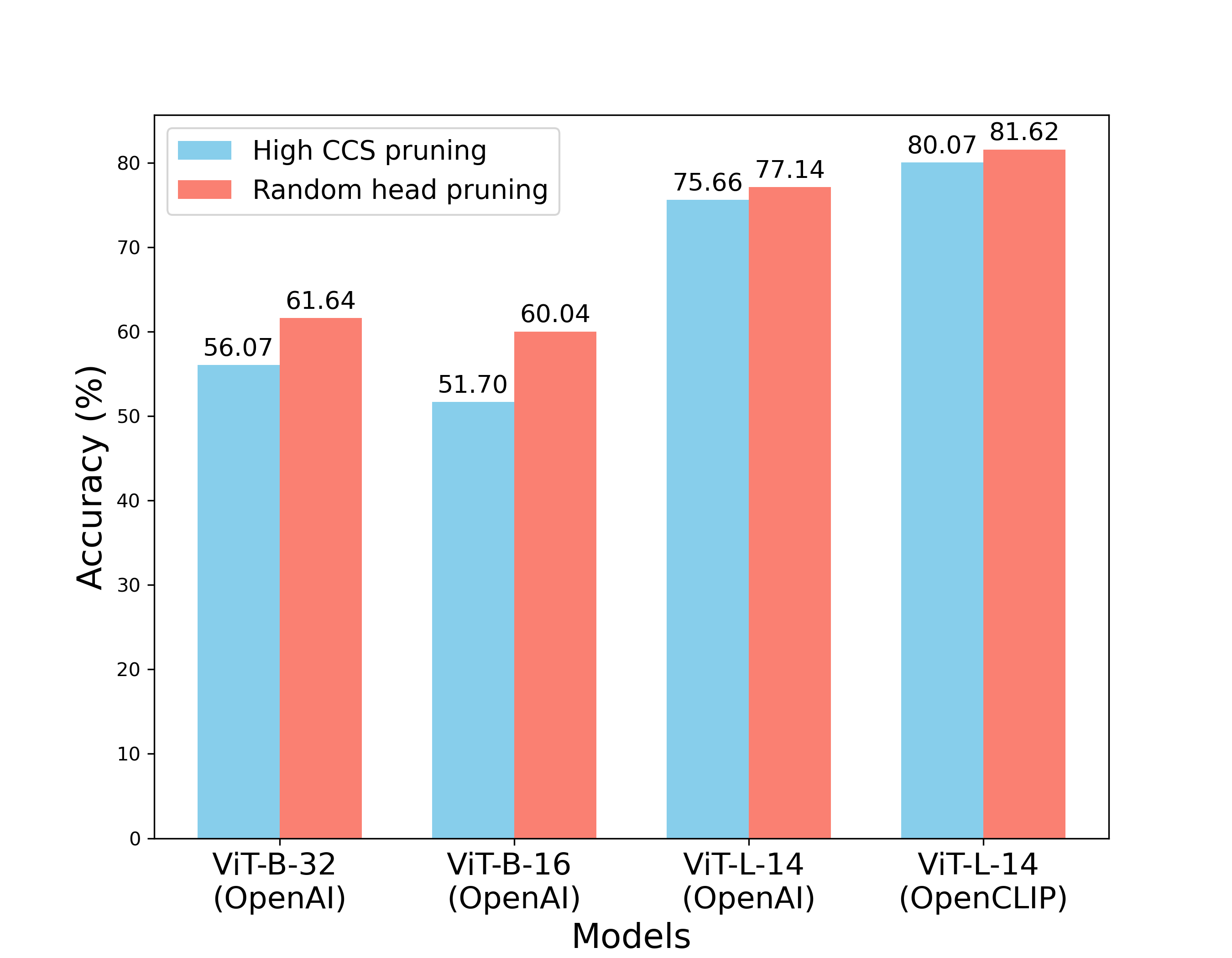}
        \caption{CIFAR-100}
        \label{fig:cifar100}
    \end{subfigure}
    \hfill
    % Third Figure
    \begin{subfigure}[b]{0.32\textwidth}
        \centering
        \includegraphics[width=\textwidth]{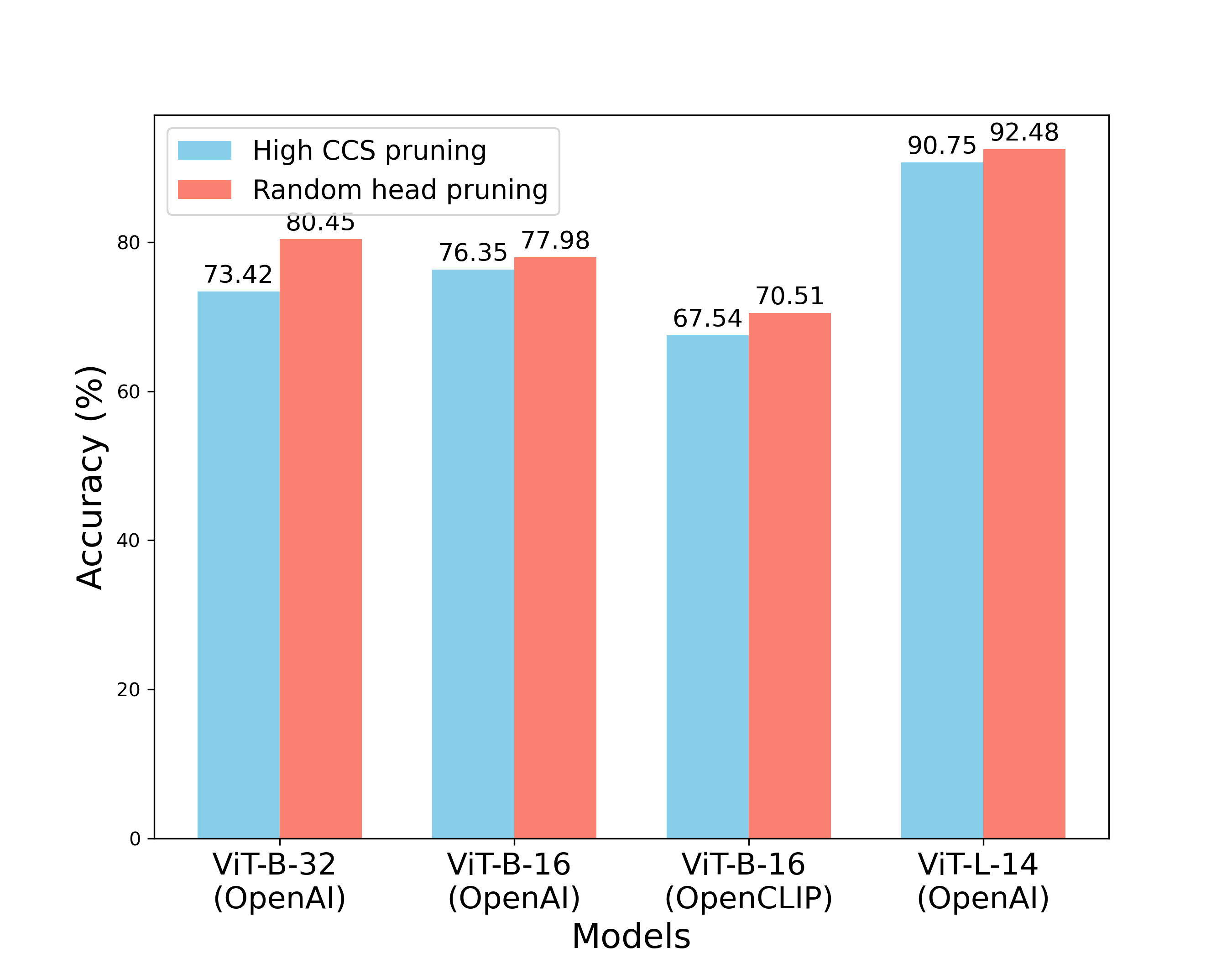}
        \caption{Food-101}
        \label{fig:food101}
    \end{subfigure}
    
    \caption{\textbf{Zero-shot performance comparison for CIFAR-10, CIFAR-100, and Food-101 datasets under different pruning strategies.} For random pruning, results are averaged across five runs.}
    \label{fig:high-random-ccs-pruning}
\end{figure*}

\begin{figure}[t]
    \centering
    \includegraphics[width=0.96\linewidth]{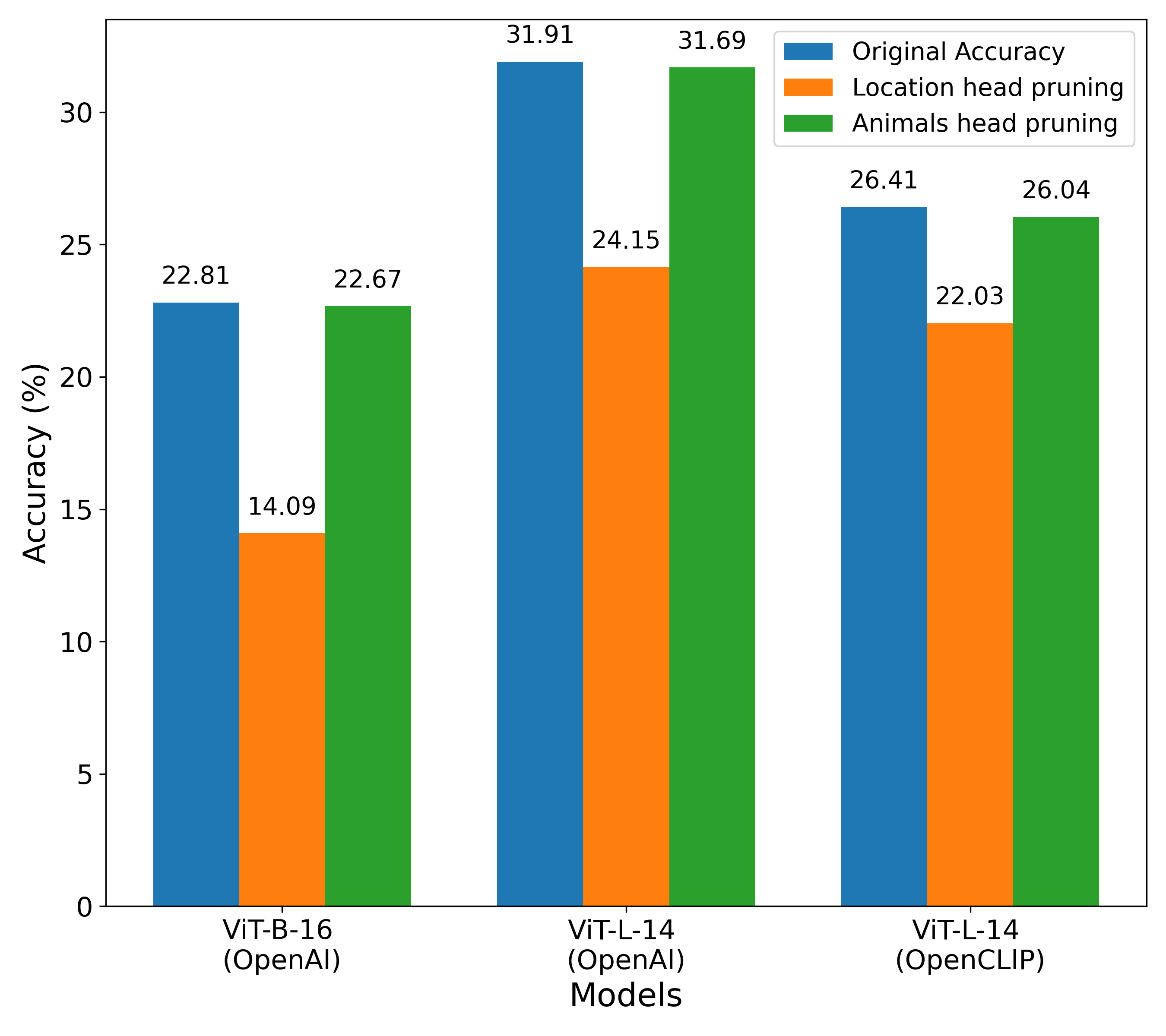}
    \caption{\textbf{Zero-shot results on Country-211 (location) dataset.}}
    \label{fig:country-locations}
\end{figure}

\begin{table*}[h]
    \scriptsize
    \centering
    % \resizebox{1\textwidth}{!} {
    \begin{tabular}{l| ccc| ccc| ccc| ccc}
        \toprule
        \multirow{4}{*}{\centering \textbf{Model}} & 
\multicolumn{3}{c}{\textbf{Country-211}} & 
\multicolumn{3}{c}{\textbf{Oxford-pets}} & 
\multicolumn{3}{c}{\textbf{ImageNet-A}} & 
\multicolumn{3}{c}{\textbf{ImageNet-R}} \\
\cmidrule(lr){2-4} \cmidrule(lr){5-7} \cmidrule(lr){8-10} \cmidrule(lr){11-13}
 & \textbf{Original} & \textbf{\shortstack{High\\CCS}} & \textbf{\shortstack{Low\\CCS}}
 & \textbf{Original} & \textbf{\shortstack{High\\CCS}} & \textbf{\shortstack{Low\\CCS}}
 & \textbf{Original} & \textbf{\shortstack{High\\CCS}} & \textbf{\shortstack{Low\\CCS}}
 & \textbf{Original} & \textbf{\shortstack{High\\CCS}} & \textbf{\shortstack{Low\\CCS}} \\
\midrule
        ViT-B-32-OpenAI & 17.16 &	11.46	& 16.3 & 50.07	& 46.66 &	48.96 & 31.49 & 20.24 & 28.72 & 69.09 & 54.47 & 64.45 \\
       ViT-B-32-datacomp & 4.43 & 4.37 & 4.37 & 26.48 & 25.98 & 25.33 & 4.96 & 4.59 & 4.65 & 34.06 & 31.6 & 32.47 \\
        \midrule
        ViT-B-16-OpenAI & 22.81 & 10.72 & 21.79 & 52.72 & 49.12 & 51.89 & 49.85 & 25.49 & 47.27 & 77.37 & 55.52 & 74.84 \\
      ViT-B-16-LAION  & 20.45 & 7.49 & 16.87 & 65.79 & 48.48 & 49.81 & 37.97 & 25.27 & 27.44 & 80.56 & 66.32 & 71.73 \\
        \midrule
        ViT-L-14-OpenAI & 31.91 & 23.21 & 30.63 & 61.79 & 62.04 & 62.08 & 70.4 & 68.15 & 69.2 & 87.87 & 86.56 & 86.97 \\
      ViT-L-14-LAION  & 26.41 & 16.38 & 25.66  & 54.1 & 56.12 & 57.16 & 53.8 & 42.44 & 52.93 & 87.12 & 82.22 & 86.94 \\
        \bottomrule
    \end{tabular}
    % }
    \caption{\textbf{Accuracy comparison of various CLIP models on Country-211, Oxford-pets, ImageNet-A and ImageNet-R datasets. The values represent original accuracy, performance after pruning high-CCS heads, and performance after pruning low-CCS heads.}}
    \label{tab:country-oxford-imagenet}
\end{table*}

\begin{table*}[h]
    \scriptsize
    \centering
    \resizebox{1\textwidth}{!} {
    \begin{tabular}{l| ccc| ccc | ccc}
        \toprule
        \multirow{2}{*}{\textbf{Model}} & \multicolumn{3}{c}{\textbf{CIFAR-100}} & \multicolumn{3}{c}{\textbf{FOOD-101}} & \multicolumn{3}{c}{\textbf{Country-211}} \\
        \cmidrule(lr){2-4} \cmidrule(lr){5-7} \cmidrule(lr){8-10}
        & \textbf{Original} & \textbf{High CCS} & \textbf{Low CCS} & \textbf{Original} & \textbf{High CCS} & \textbf{Low CCS} & \textbf{Original} & \textbf{High CCS} & \textbf{Low CCS} \\
        \midrule
        ViT-B-32-OpenAI & 65.08 & 62.39 & 58.58 & 84.01 & 82.12 & 73.24 & 17.16 & 16.3 & 12.23 \\
ViT-B-32-datacomp & 54.95 & 53.72 & 53.77 & 41.66 & 40.77 & 39.31 & 4.43 & 4.37 & 4.41 \\
\midrule
ViT-B-16-OpenAI & 68.22 & 65.38 & 56.17 & 88.73 & 87.36 & 83.23 & 22.81 & 21.79 & 21.74 \\
ViT-B-16-LAION  & 76.92 & 72.51 & 72.73 & 86.63	& 81.4 & 80.51 & 20.45 & 16.87 & 14.39 \\
\midrule
ViT-L-14-OpenAI & 78.28 & 77.55 & 73 & 93.07 & 92.79 & 90.15 & 31.91 & 30.63 & 25.38 \\
ViT-L-14-LAION & 83.37 & 83.25 & 83.35 & 91.02 & 90.35 & 90.25 & 26.41 & 25.66 & 25.43 \\
        \bottomrule
    \end{tabular}
    }
    \caption{\textbf{Accuracy comparison of various CLIP models on CIFAR-100, FOOD-101 and Country-211 datasets. The values represent original accuracy, performance after pruning equal number of high-CCS and low-CCS heads.}}
    \label{tab:cifar-food-pruning-equal}
\end{table*}

\section{Results and Discussion} 
\subsection{Interpretable CLIP Models: The Role of CCS.} \label{sec:ccs_pruning}
In this section we examine the role of the Concept Consistency Score (CCS) in revealing CLIP's decision-making process, focusing on the question: \textit{How does CCS provide deeper insights into the functional role of individual attention heads in influencing downstream tasks?} To explore this, we perform a soft-pruning analysis by zeroing out attention weights of heads with extreme CCS values—specifically, high CCS (CCS = 5) and low CCS (CCS $\le$ 1). This approach disables selected heads without modifying the model architecture. As shown in Table ~\ref{tab:cifar-food}, pruning high-CCS heads consistently causes significant drops in zero-shot classification performance across CIFAR-10, CIFAR-100 and FOOD-101 while pruning low-CCS heads has a minimal effect. This performance gap demonstrates that CCS effectively identifies heads encoding critical, concept-aligned information, making it a reliable tool for interpreting CLIP's internal decision-making mechanisms.

We further observe notable variations in pruning sensitivity across model architectures. ViT-B-16 models suffer the most from high-CCS head pruning, implying a reliance on a smaller number of specialized heads. In contrast, ViT-L-14 models show greater resilience, suggesting more distributed representations. Among smaller models, OpenAI-trained models experience larger performance drops than OpenCLIP models when high-CCS heads are pruned. However, in larger models like ViT-L-14, OpenCLIP variants show a slightly higher degradation. These patterns reveal that CCS not only identifies functionally important heads but also captures model-specific and training-specific differences in how conceptual knowledge is organized and utilized within CLIP architectures.

\subsection{Pruning equal number of high CCS, low CCS and random heads}
% \phillip{it seems like this section should be merged with the prior one}
In the previous section, we showed that attention heads with high Concept Consistency Scores (CCS) are crucial to CLIP’s performance. To validate whether these heads are truly more important than others, we perform a controlled comparison against random pruning. Specifically, we randomly prune the same number of attention heads—excluding high-CCS heads—and repeat this across five seeds, averaging the results. As illustrated in Figure ~\ref{fig:high-random-ccs-pruning}, pruning high-CCS heads consistently causes a significantly larger drop in zero-shot accuracy compared to random pruning across datasets and model variants. In contrast, random pruning results in only minor performance degradation, highlighting the functional importance of high-CCS heads. Interestingly, we also find that larger CLIP models show a smaller performance gap between high-CCS and random pruning, suggesting that larger architectures may be more robust due to greater redundancy or more distributed representations. 

Similarly, we conducted experiments where we pruned an equal number of high and low CCS attention heads across multiple datasets (CIFAR-100, FOOD-101, Country-211). Results are shown in the table ~\ref{tab:cifar-food-pruning-equal}. From the table, we observe that pruning high-CCS heads leads to a substantially larger performance drop, even when the number of pruned heads is held constant. This effectively rules out the explanation that the observed degradation is merely due to pruning more heads in the high-CCS condition. Taken together, these findings support CCS as a reliable and interpretable metric for identifying concept-relevant heads and offer deeper insights into how CLIP organizes conceptual information.

\begin{figure}
    \centering
    \includegraphics[width=\linewidth]{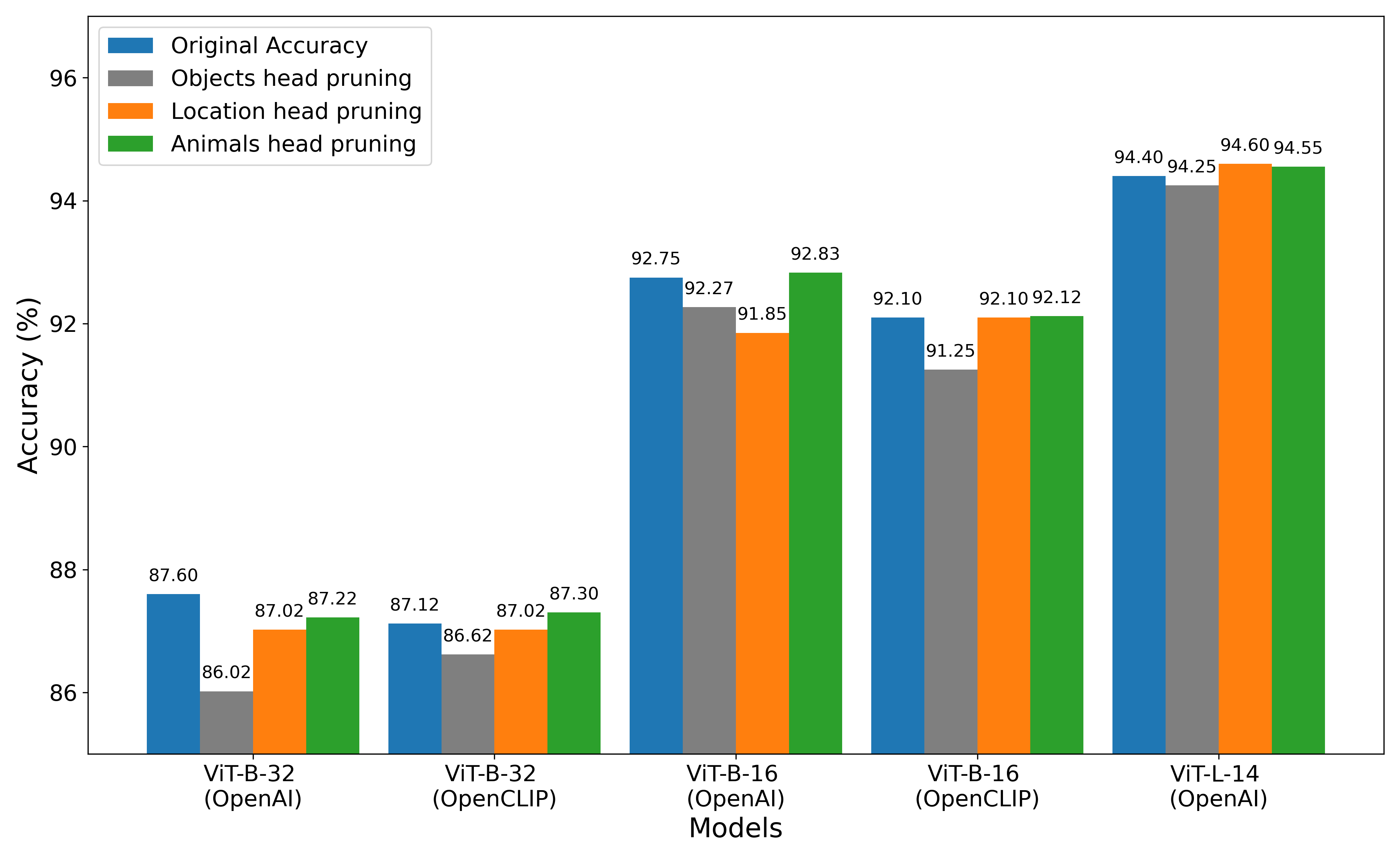}
    \caption{\textbf{Zero-shot results on CIFAR-10 (Objects) dataset.}}
    \label{fig:cifar-objects}
\end{figure}

\begin{figure*}[ht]
    \centering
    % First Figure
    \begin{subfigure}[b]{0.33\textwidth}
        \centering
        \includegraphics[width=\textwidth]{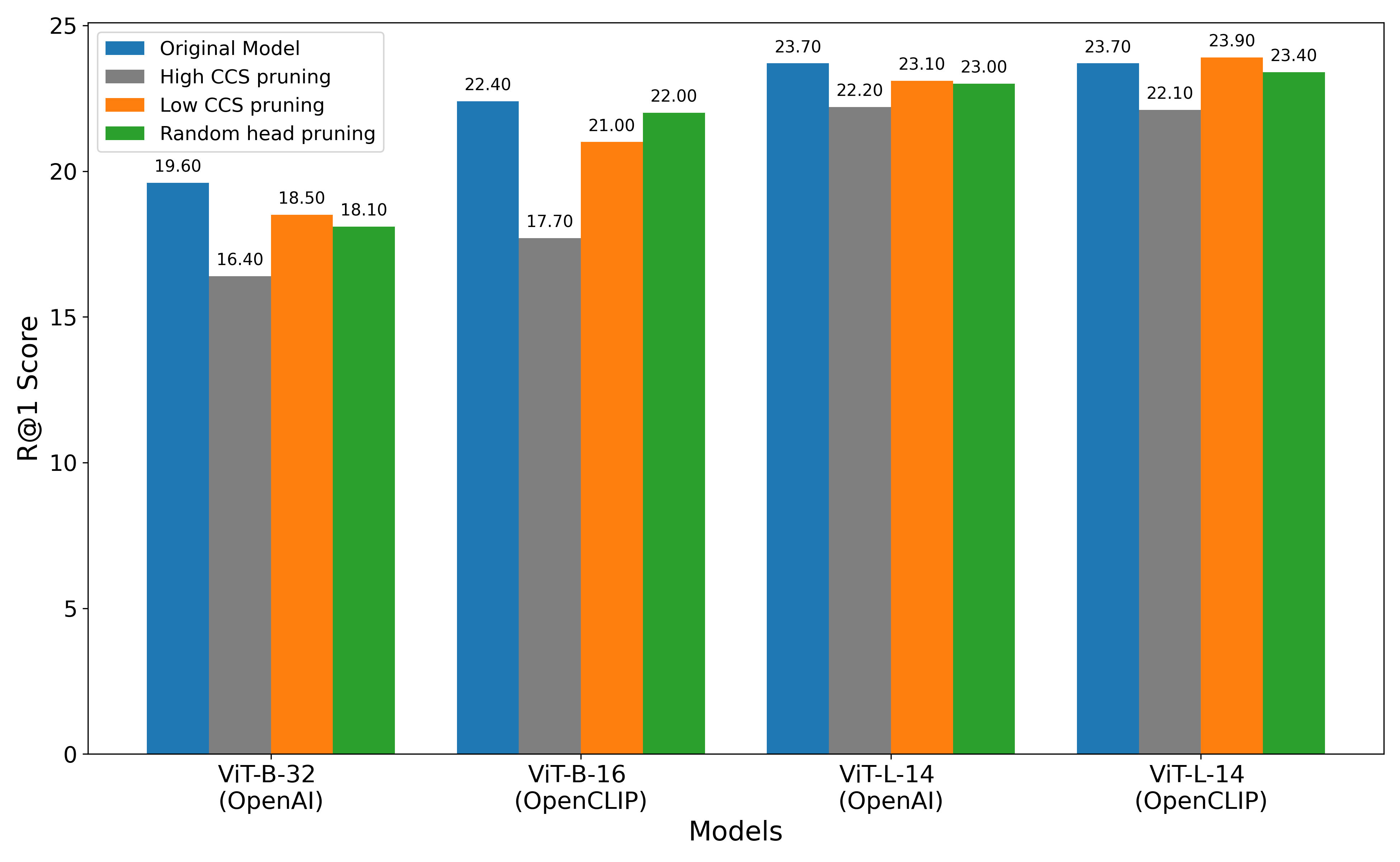}
        \caption{MSRVTT}
        \label{fig:msrvtt}
    \end{subfigure}
    \hfill
    % Second Figure
    \begin{subfigure}[b]{0.33\textwidth}
        \centering
        \includegraphics[width=\textwidth]{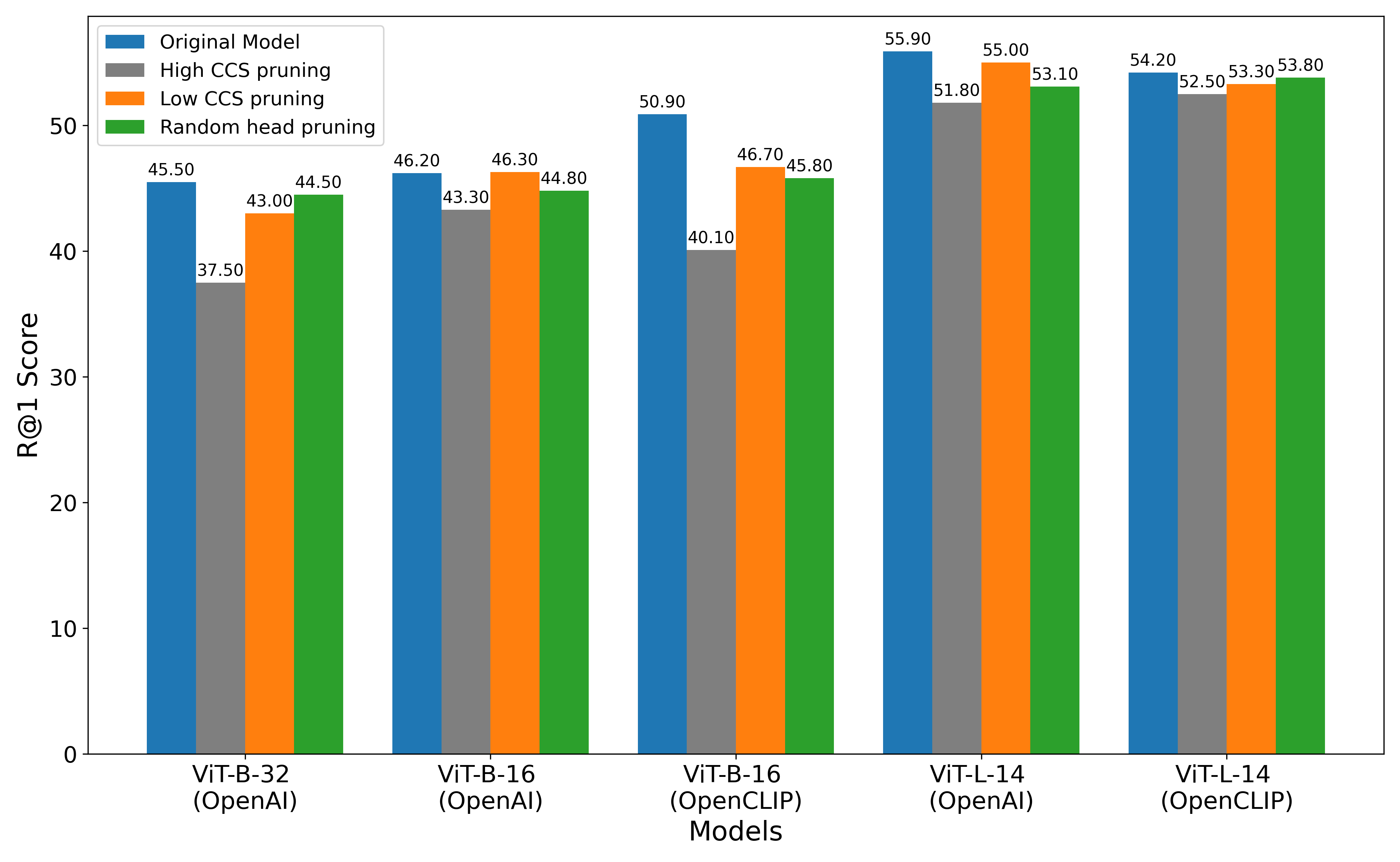}
        \caption{MSVD}
        \label{fig:msvd}
    \end{subfigure}
    \hfill
    % Third Figure
    \begin{subfigure}[b]{0.32\textwidth}
        \centering
        \includegraphics[width=\textwidth]{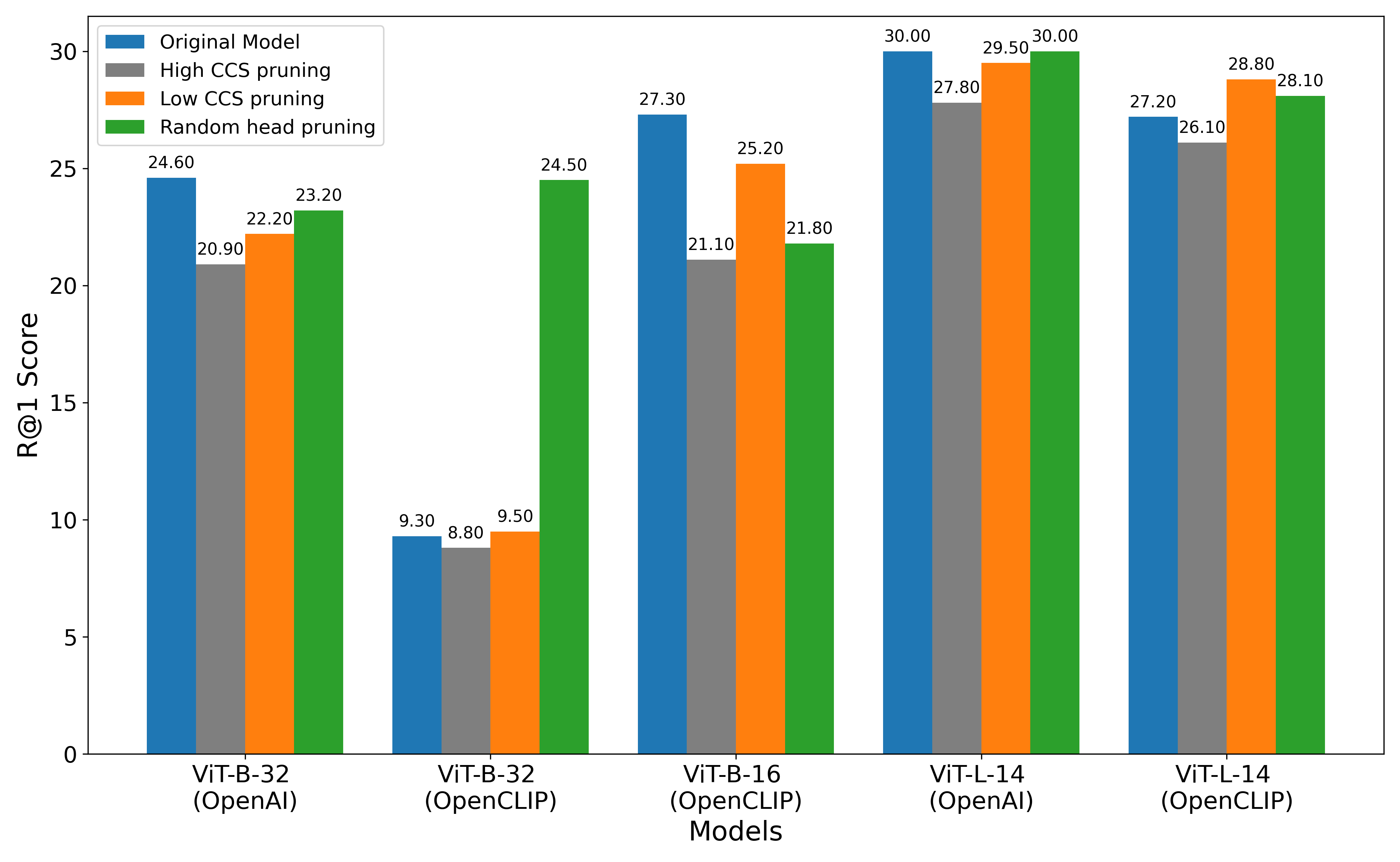}
        \caption{DIDEMO}
        \label{fig:didemo}
    \end{subfigure}
    \caption{\textbf{Zero-shot performance comparison of unpruned (original) model, pruning high CSS, low CSS and random heads on video retrieval task.}}
    \label{fig:video-ret-ccs}
\end{figure*}

\subsection{High CCS heads are crucial for out-of-domain (OOD) detection}
While our earlier experiments primarily focused on in-domain datasets such as CIFAR-10 and CIFAR-100 to validate the Concept Consistency Score (CCS), understanding model behavior under out-of-domain (OOD) conditions is a critical step toward evaluating models' robustness. Table ~\ref{tab:country-oxford-imagenet} demonstrates the results on ImageNet-A and ImageNet-R datasets respectively. From the table, we observe that pruning heads with high CCS scores leads to a substantial degradation in model performance, underscoring the critical role these heads play in the model's decision-making process. Notably, the ViT-B-16-OpenAI model exhibits the most pronounced drop in performance upon pruning high CCS heads, suggesting that this model relies heavily on a smaller set of concept-specific heads for robust feature representation consistent with the observations previously. These results demonstrate that CCS is a powerful metric for identifying attention heads that encode essential, generalizable concepts in CLIP models.

\subsection{High CCS heads are crucial for concept-specific tasks.}
To investigate the functional role of high Concept Consistency Score (CCS) heads, we conduct concept-specific pruning experiments. In these experiments, we prune heads with high CCS scores corresponding to a target concept (e.g., locations) and evaluate the model's performance on tasks aligned with that concept, such as location classification. In contrast, we also prune heads associated with unrelated concepts (e.g., animals) and assess the resulting impact on task performance. Our results indicate that pruning high CCS heads leads to a significant drop in task performance, validating that these heads encode essential concept-relevant information. For instance, in the ViT-B-16 model, pruning location heads results in a substantial decrease in location classification accuracy from 22.81\% to 14.09\%, as shown in Figure~\ref{fig:country-locations}. Conversely, pruning heads corresponding to unrelated concepts has little effect on performance, demonstrating the concept-specific nature of high CCS heads, as illustrated in Figure~\ref{fig:cifar-objects}.

In more general classification tasks, object-related heads consistently exhibit a greater impact on performance than location or color heads. For example, in the ViT-B-32 model, pruning object-related heads leads to a more noticeable accuracy drop (from 87.6\% to 86.02\%) compared to pruning location or color heads, which result in smaller reductions (87.02\% and 87.22\%, respectively). This underscores the greater importance of object-related features in vision tasks. Larger models, such as ViT-L-14, demonstrate a more robust performance to pruning, with smaller accuracy drops when pruning concept-specific heads, suggesting that these models employ more distributed and redundant representations. For instance, pruning object-related heads in ViT-L-14 reduces accuracy only marginally, from 92.1\% to 91.25\%, with negligible effects from pruning location and color heads. These results not only confirm the effectiveness of CCS as an interpretability tool but also show that high CCS heads are critical for concept-aligned tasks and provide significant insights into how concepts are represented within CLIP-like models.
%Moreover, the reduced impact of pruning in larger models highlights differences in the representation capacity and redundancy across different model architectures.

\subsection{Impact of CCS pruning on zero-shot video retrieval.}
To further assess the importance of high CCS heads for downstream tasks, we conducted zero-shot video retrieval experiments on three popular datasets (MSRVTT, MSVD, and DIDEMO) under different pruning strategies. The results in Figure ~\ref{fig:video-ret-ccs} show that pruning high CCS (Concept Consistency Score) heads consistently leads to a substantial drop in performance across all datasets, demonstrating their critical role in preserving CLIP's retrieval capabilities. For instance, on MSRVTT and MSVD, high CCS pruning significantly underperforms compared to low CCS and random head pruning, which show much milder performance degradation. Interestingly, low CCS and random head pruning maintain performance much closer to the original unpruned model, indicating that not all attention heads contribute equally to model competence. This consistent trend across datasets highlights that heads with high CCS scores are essential for encoding concept-aligned information necessary for accurate zero-shot video retrieval.

\begin{table}
    \tiny
    \centering
    %\resizebox{1\textwidth}{!} {
    \begin{tabular}{l| ccc | ccc}
        \toprule
        \multirow{4}{*}{\centering \textbf{Model}} & \multicolumn{3}{c}{\textbf{Race ($\downarrow$)}} & \multicolumn{3}{c}{\textbf{Gender ($\downarrow$)}} \\
        \cmidrule(lr){2-4} \cmidrule(lr){5-7}
        & \textbf{Original} & \textbf{\shortstack{High\\CCS}} & 
        \textbf{\shortstack{Low\\CCS}} &
        \textbf{Original} & \textbf{\shortstack{High\\CCS}} &
        \textbf{\shortstack{Low\\CCS}}\\
        \midrule
        ViT-B-32-OpenAI & 61.8 & 60.5 & 65.0 & 41.2 & 41.1 & 42.5 \\
ViT-B-32-datacomp & 49.2 & 48.3 & 49.7 & 21.7 & 21 & 21.6 \\
\midrule
ViT-B-16-OpenAI & 64.6 & 55.6 & 66.6 & 40.2 & 38.2 & 40.5 \\
ViT-B-16-LAION & 59.2 & 56.7 & 56.6 & 43.6 & 43.1 & 43.8 \\
\midrule
ViT-L-14-OpenAI & 59.3 & 59.8 & 59.9 & 34.7 & 32.2 & 35.2 \\
ViT-L-14-LAION & 61.9 & 55.6 & 59.7 & 43 & 39.2 & 43.9 \\
        \bottomrule
    \end{tabular}
   % }
    \caption{\textbf{Comparison of original and high-CCS pruning on FairFace dataset for race and gender.}. We used MaxSkew (K=900) as the metric.}
    \label{tab:fairface-fairness}
\end{table}

\begin{table}
    \tiny
    \centering
    %\resizebox{1\textwidth}{!} {
    \begin{tabular}{l| ccc| ccc}
        \toprule
        \multirow{4}{*}{\centering \textbf{Model}} & \multicolumn{3}{c}{\textbf{Race ($\downarrow$)}} & \multicolumn{3}{c}{\textbf{Gender ($\downarrow$)}} \\
        \cmidrule(lr){2-4} \cmidrule(lr){5-7}
        & \textbf{Original} & \textbf{\shortstack{High\\CCS}} & \textbf{\shortstack{Low\\CCS}} & \textbf{Original} & \textbf{\shortstack{High\\CCS}}  & \textbf{\shortstack{Low\\CCS}}\\
        \midrule
        ViT-B-32-OpenAI & 4.1 & 1.2 & 2.4 & 3.7 & 2.4 & 1.2\\
ViT-B-16-OpenAI & 0.8 & 2.0 & 1.2 & 2.4 & 1.2 & 2.1  \\
ViT-L-14-OpenAI & 2.0 & 0.8 & 2.0 & 2.4	& 2.0 & 1.6\\
        \bottomrule
    \end{tabular}
   % }
    \caption{\textbf{Comparison of original and high-CCS soft-pruning on SocialCounterFactuals dataset for race and gender.}. We used MaxSkew (K=12 for race, K=4 for gender) as the metric.}
    \label{tab:soco-fairness}
\end{table}

\subsection{CLIP's high-CCS heads encode features that drive social biases.}
Previously, we established that high-CCS heads in CLIP models are crucial for image and video tasks and pruning them leads to significant drop in performance. Now, we investigate if these high CCS heads learn spurious features leading to social biases. For this, we perform soft pruning experiment on FairFace and SocialCounterFactuals datasets. Here given neutral text prompts of 104 occupations\footnote{List of occupations and prompts can be foudn in Appendix}, we measure MaxSkew across race and gender in the datasets. Tables ~\ref{tab:fairface-fairness} and ~\ref{tab:soco-fairness} show the results on FairFace and SocialCounterFactuals datasets respectively. 

On the FairFace dataset, pruning high-CCS heads consistently reduces the MaxSkew values for both race and gender across all models. For example, in the ViT-B-16-OpenAI model, pruning high-CCS heads drops the race MaxSkew from 64.6 to 55.6 and the gender MaxSkew from 40.2 to 38.2. Similar reductions are observed across all ViT-B and ViT-L variants. These drops, although modest in some cases, indicate a consistent trend: high-CCS heads are contributing disproportionately to skewed model predictions. The effect is even more evident on the SocialCounterfactuals dataset, where MaxSkew values drop substantially upon pruning high-CCS heads. For instance, in ViT-B-32-OpenAI, the race MaxSkew falls from 4.1 to 1.2, and gender MaxSkew from 3.7 to 2.4. Similar reductions occur for other ViT variants, with some pruned models showing more than 50\% decrease in bias.

These results reveal a fundamental paradox at the heart of CLIP models: high-CCS heads, while critical for strong performance in tasks such as classification, retrieval, and concept alignment, are also the primary contributors to social bias. Pruning these heads leads to a notable reduction in model bias, as shown in our experiments, but also comes at the cost of reduced performance—a clear trade-off between fairness and utility. This performance-bias paradox underscores the complex role of high-CCS heads: they are both enablers of semantic understanding and carriers of learned stereotypes. The CCS metric, in this context, provides a valuable lens for navigating this tension. It not only aids in interpreting model behavior but also offers a lightweight intervention—soft-pruning—that mitigates bias without requiring expensive fine-tuning.
\section{Conclusion}
In this work, we proposed Concept Consistency Score (CCS), a novel interpretability metric that quantifies how consistently individual attention heads in CLIP-like models align with semantically meaningful concepts. Through extensive soft-pruning experiments, we demonstrated that heads with high CCS are essential for maintaining model performance, as their removal leads to substantial performance drops compared to pruning random or low CCS heads. Our findings further highlight that high CCS heads are not only critical for standard vision-language tasks but also play a central role in out-of-domain detection and concept-specific reasoning. Moreover, experiments on video retrieval tasks reveal that high CCS heads are crucial for capturing temporal and cross-modal relationships, underscoring their broad utility in multimodal understanding. In addition, we demonstrated that high-CCS heads learn spurious correlations leading to social biases and pruning them mitigates that harmful behaviour without the need for further finetuning. Thus, CCS provides an wholistic view of interpretability proving the paradox performance vs social biases in CLIP.

\section{Limitations}
In this work, we experimented primarily on CLIP models. Although CCS metric established the fundamental paradox of performance vs social biases we haven't proved for other vision language models. Hence, we leave extending for more vision language models for future work.  
Another limitation is the use of LLM models for concept labelling and judging which requires robust manual verification to limit any inconsistencies. Hence, scaling our work to much bigger models with more layers and heads can be a limitation.

\bibliography{custom}
\clearpage
\appendix

\begin{table*}[t]
    \centering
    \resizebox{1\textwidth}{!}
    {
    \begin{tabular}{ccccccccccc}
    \toprule
     \textbf{Model} & \textbf{Model size} & \textbf{Patch size} & \textbf{Pre-training data} & \textbf{CCS@0 } & \textbf{CCS@1 } & \textbf{CCS@2} & \textbf{CCS@3 } & \textbf{CCS@4 } & \textbf{CCS@5} \\
    \midrule
    CLIP & B & 32 & OpenAI-400M  & 0.021 & 0.062 & 0.167 & 0.271 & 0.312 & 0.167 \\
    CLIP & B   & 32 & OpenCLIP-datacomp & 0.104 & 0.062 & 0.208 & 0.189 & 0.208 & 0.229 \\
    CLIP & B  & 16 & OpenAI-400M & 0.021 & 0.062 & 0.125 & 0.292 & 0.292 & 0.208   \\
CLIP & B   & 16 & OpenCLIP-LAION2B & 0.062 & 0.062 & 0.105 & 0.25 & 0.25 & 0.271   \\
CLIP & L   & 14 & OpenAI-400M & 0.062 & 0.109 & 0.172 & 0.204 & 0.203 & 0.25   \\
CLIP & L   & 14 & OpenCLIP-LAION2B & 0.016 & 0.031 & 0.109 & 0.219 & 0.297 & 0.328   \\
    \bottomrule
    \end{tabular}
    }
    \caption{\textbf{Concept Consistency Score (CCS) for CLIP models}.} 
    \label{tab:ccs-score}
\end{table*}

\section{Concept Consistency Scores (CCS) for CLIP models.}
We measure $CCS@K$ for all values of $K$ i.e $K \in [0, 5]$.  Table~\ref{tab:ccs-score} presents the Concept Consistency Score (CCS) distribution across various CLIP models, categorized by architecture size, patch size, and pre-training data. Several noteworthy trends emerge from this analysis. First, models pre-trained on larger and more diverse datasets (e.g., OpenCLIP-LAION2B) tend to exhibit a higher proportion of heads with CCS@5, indicating that a greater number of transformer heads are aligned with semantically meaningful concepts. For instance, the ViT-L-14 model trained on LAION2B shows the highest CCS@5 score of 0.328, suggesting that approximately 32.8\% of heads are consistently associated with a single concept, reflecting strong concept alignment in these models.

Second, smaller models such as ViT-B-32 trained on OpenAI-400M demonstrate a significantly lower CCS@5 score (0.167) and a higher proportion of heads with lower CCS values (e.g., CCS@0 = 0.021), indicating weaker alignment of heads to consistent concepts. This observation implies that larger models with richer pre-training data are better at learning concept-specific representations, a key requirement for robust and interpretable multimodal reasoning.

Interestingly, when comparing models with the same architecture but different pre-training corpora, such as ViT-B-32 (OpenAI-400M vs. OpenCLIP-datacomp), we observe a higher CCS@5 score for datacomp (0.229) than OpenAI-400M (0.167), suggesting that dataset composition significantly affects the emergence of interpretable heads.

Moreover, progressive increases in CCS from CCS@0 to CCS@5 show how concept alignment varies within each model. For instance, while ViT-L-14 (OpenCLIP-LAION2B) has a low CCS@0 of 0.016, it steadily increases to a high CCS@5 of 0.328, suggesting that although a few heads are poorly aligned, a substantial fraction are highly consistent in capturing specific concepts.

In summary, these results demonstrate that the CCS metric effectively captures differences in conceptual alignment across models of varying size and pre-training datasets. Models with larger capacities and richer pre-training datasets tend to exhibit higher concept consistency, offering better interpretability and potentially stronger generalization abilities. This analysis underscores the value of CCS as a diagnostic tool for evaluating and comparing the internal conceptual representations learned by CLIP-like models.

\begin{table*}[h]
\centering
\begin{tabular}{l c}
\toprule
\textbf{Model} & \textbf{Concentrated Concept Ratio (CCR)} \\
\midrule
ViT-B-32-OpenAI      & 0.250 \\
ViT-B-32-datacomp    & 0.273 \\
ViT-B-16-OpenAI      & 0.200 \\
ViT-B-16-LAION       & 0.231 \\
ViT-L-14-OpenAI      & 0.125 \\
ViT-B-14-LAION       & 0.143 \\
\bottomrule
\end{tabular}
\caption{\textbf{Concentrated Concept Ratio (CCR) across different CLIP model variants}. Lower CCR in larger models (e.g., ViT-L-14) suggests more distributed concept representations.}
\label{tab:ccr_values}
\end{table*}

\section{Redundancy and Functional Duplication Analysis.}
Previously in section ~\ref{sec:ccs_pruning}, we observed that larger models degrade less when high-CCS (Concept Contribution Score) attention heads are pruned. This raises the question of whether larger models exhibit greater redundancy—i.e., whether multiple heads are attending to the same concepts, providing robustness against pruning.

To quantify this, we introduce the Concentrated Concept Ratio (CCR), a metric that captures the extent to which concepts are redundantly represented across multiple high-CCS heads. 

\begin{equation}
   CCR = \frac{C_{multi}}{H_{high}}
\end{equation}

where $H_{high}$ be the set of attention heads with high CCS, $C_{multi}$ be the number of unique concepts that are concentrated on more than one head within $H_{high}$. A higher CCR implies greater redundancy—i.e., the same concepts are captured by multiple heads. Conversely, a lower CCR suggests a more distributed and unique representation of concepts across heads.

Table ~\ref{tab:ccr_values} presents CCR values across models of varying sizes. Interestingly, larger models (e.g., ViT-L variants) show lower CCR, indicating that they tend to distribute conceptual information more evenly across heads. This aligns with the empirical finding that these models are more resilient to pruning of high-CCS heads.

To further examine whether redundancy exists across layers (functional duplication), we perform a layer-wise CCR analysis restricted to the last four transformer layers of each model. This focus is motivated by prior work  ~\cite{gandelsmaninterpreting}, which shows that only the final four layers significantly affect model outputs. CCR is computed independently for each of the last four layers. However, the analysis reveals no consistent trend indicating that deeper layers exhibit greater concept duplication. This suggests that redundancy across layers is not a dominant factor and that the observed robustness in larger models is more likely due to distributed head-level representations rather than functional duplication across layers.

\begin{table*}
\centering
\small
% \resizebox{\textwidth}{!}{%
\begin{tabular}{p{11.25cm}}
\toprule
\textbf{ViT-B-32-OpenAI} \\
L8.H11 (Descriptive), L9.H2 (Objects), L9.H3 (Descriptions), L10.H8 (Locations), L11.H1 (Objects), L11.H5 (Colors), L11.H7 (Objects), L11.H9 (Locations) \\
\midrule
\textbf{ViT-B-32-datacamp} \\
L8.H1 (Objects), L8.H3 (Subjects), L8.H10 (Objects), L9.H3 (Subjects), L9.H10 (Objects), L10.H7 (Locations), L10.H11 (Objects), L11.H3 (Colors), L11.H4 (Colors), L11.H9 (Colors), L11.H10 (Objects) \\
\midrule
\textbf{ViT-B-16-OpenAI} \\
L8.H5 (Visual), L8.H8 (Visual), L10.H5 (Subjects), L10.H7 (Settings), L11.H0 (Creative), L11.H3 (Settings), L11.H4 (Stylistic), L11.H6 (Locations), L11.H7 (Colors), L11.H11 (Animals) \\
\midrule
\textbf{ViT-B-16-LAION} \\
L8.H6 (Descriptions), L8.H7 (Descriptions), L9.H0 (Themes), L9.H1 (Aesthetics), L9.H3 (Descriptive), L10.H5 (Artwork), L10.H10 (Locations), L11.H0 (Locations), L11.H2 (Descriptions), L11.H6 (Locations), L11.H7 (Objects), L11.H8 (Objects), L11.H10 (Colors) \\
\midrule
\textbf{ViT-L-14-OpenAI} \\
L20.H2 (Locations), L20.H12 (Descriptions), L21.H0 (Locations), L21.H1 (Locations), L21.H8 (Expressions), L21.H13 (Locations), L21.H15 (Locations), L22.H1 (Objects), L22.H2 (Locations), L22.H5 (Locations), L22.H9 (Subjects), L22.H13 (Animals), L22.H15 (Locations), L23.H4 (Objects), L23.H10 (Locations), L23.H11 (Colors) \\
\midrule
\textbf{ViT-L-14-LAION} \\
L20.H4 (Subjects), L20.H14 (Descriptions), L21.H0 (Colors), L21.H1 (Locations), L21.H5 (Descriptive), L21.H9 (Colors), L21.H11 (Locations), L22.H0 (Patterns), L22.H1 (Shapes), L22.H3 (Objects), L22.H5 (Visual), L22.H6 (Animals), L22.H8 (Letters), L22.H10 (Colors), L22.H12 (Landscapes), L22.H13 (Locations), L23.H4 (People), L23.H5 (Nature), L23.H6 (Locations), L23.H8 (Colors), L23.H9 (Descriptive) \\
\bottomrule
\end{tabular}
% }
\caption{\textbf{Full List of high-CCS heads of all CLIP models.}}
\label{tab:high-ccs}
\end{table*}

\begin{table*}
\centering
\small
% \resizebox{\textwidth}{!}{%
\begin{tabular}{p{11.25cm}}
\toprule
\textbf{ViT-B-32-OpenAI} \\
L8.H1 (Artistic), L8.H2 (Objects), L8.H6 (Photography), L8.H9 (Styles), L8.H10 (Perspective), L9.H1 (Subjects), L9.H11 (Settings), L10.H0 (Objects), L10.H3 (Locations), L10.H7 (Locations), L11.H6 (Descriptions), L11.H10 (Locations), L11.H11 (Locations) \\
\midrule
\textbf{ViT-B-32-datacamp} \\
L8.H0 (Environments), L8.H7 (Creativity), L9.H6 (Colors), L10.H5 (Art), L10.H6 (Descriptions), L10.H8 (Locations), L10.H9 (Descriptions), L11.H2 (Subjects), L11.H8 (Qualities) \\
\midrule
\textbf{ViT-B-16-OpenAI} \\
L8.H1 (Artistic), L8.H2 (Photography), L8.H4 (Styles), L8.H6 (Artwork), L8.H7 (Photography), L8.H9 (Light), L9.H4 (Photography), L9.H6 (Artforms), L9.H10 (Elements), L10.H3 (Locations), L10.H8 (Colors), L10.H9 (Artwork), L11.H5 (Objects), L11.H8 (Effects) \\
\midrule
\textbf{ViT-B-16-LAION} \\
L8.H0 (Locations), L8.H8 (Text), L8.H9 (Photography), L9.H7 (Artistic), L9.H8 (Settings), L9.H11 (Descriptions), L10.H2 (Nature), L10.H3 (Location), L10.H7 (Expressions), L11.H3 (Settings), L11.H9 (Numbers), L11.H11 (Letters) \\
\midrule
\textbf{ViT-L-14-OpenAI} \\
L20.H0 (Locations), L20.H3 (Locations), L20.H7 (Communication), L20.H8 (Vehicles), L20.H10 (Locations), L21.H4 (Photography), L21.H6 (People), L21.H10 (Locations), L22.H3 (Countries), L22.H12 (Professions), L23.H3 (Patterns), L23.H9 (Creativity), L23.H15 (Visual) \\
\midrule
\textbf{ViT-L-14-LAION} \\
L20.H0 (Locations), L20.H1 (Locations), L20.H2 (Locations), L20.H8 (Locations), L20.H9 (Locations), L20.H11 (Aesthetics), L20.H15 (Descriptions), L21.H12 (Photography), L21.H14 (Locations), L22.H9 (Activities), L22.H14 (Colors), L22.H15 (Emotions), L23.H0 (Materials), L23.H3 (Settings) \\
\bottomrule
\end{tabular}
% }
\caption{\textbf{Full List of medium-CCS heads of all CLIP models.}}
\label{tab:medium-ccs}
\end{table*}

\begin{table*}
\centering
\small
% \resizebox{\textwidth}{!}{%
\begin{tabular}{p{11.25cm}}
\toprule
\textbf{ViT-B-32-OpenAI} \\
L8.H5 (Patterns), L9.H9 (Ambiance), L11.H0 (Diverse), L11.H8 (Word) \\
\midrule
\textbf{ViT-B-32-datacamp} \\
L8.H2 (Images), L8.H4 (Varied), L8.H9 (Varied), L9.H4 (Variety), L9.H5 (Professions), L11.H0 (Diverse), L11.H1 (Varied), L11.H11 (Settings) \\
\midrule
\textbf{ViT-B-16-OpenAI} \\
L8.H0 (Diversity), L9.H3 (Locations), L10.H6 (Body parts), L11.H2 (Perspective) \\
\midrule
\textbf{ViT-B-16-LAION} \\
L8.H4 (Variety), L8.H5 (Varied), L8.H10 (Diverse), L9.H2 (Textures), L10.H6 (Photography), L10.H8 (Traits) \\
\midrule
\textbf{ViT-L-14-OpenAI} \\
L20.H1 (Diverse), L20.H4 (Diversity), L20.H6 (Items), L20.H15 (Diverse), L21.H2 (Diversity), L21.H3 (Diverse), L22.H0 (Occupations), L22.H4 (Settings), L22.H6 (Weather), L22.H14 (Items), L23.H5 (Diversity)\\
\midrule
\textbf{ViT-L-14-LAION} \\
L20.H13 (Photography), L21.H6 (Professions), L23.H1 (Diverse) \\
\bottomrule
\end{tabular}
% }
\caption{\textbf{Full List of low-CCS heads of all CLIP models.}}
\label{tab:low-ccs}
\end{table*}

\begin{table*}
\centering
\small
% \resizebox{\textwidth}{!}{%
\begin{tabular}{p{11.25cm}}
\toprule
\textbf{Occupations} \\
\midrule
biologist, composer, economist, mathematician, model, poet, reporter, zoologist, artist, coach, athlete, audiologist, judge, musician, therapist, banker, ceo, consultant, prisoner, assistant, boxer, commander, librarian, nutritionist, realtor, supervisor, architect, priest, guard, magician, producer, teacher, lawyer, paramedic, researcher, physicist, pediatrician, surveyor, laborer, statistician, dietitian, sailor, tailor, attorney, army, manager, baker, recruiter, clerk, entrepreneur, sheriff, policeman, businessperson, chief, scientist, carpenter, florist, optician, salesperson, umpire, painter, guitarist, broker, pensioner, soldier, astronaut, dj, driver, engineer, cleaner, cook, housekeeper, swimmer, janitor, pilot, mover, handyman, firefighter, accountant, physician, farmer, bricklayer, photographer, surgeon, dentist, pianist, hairdresser, receptionist, waiter, butcher, videographer, cashier, technician, chemist, blacksmith, dancer, doctor, nurse, mechanic, chef, plumber, bartender, pharmacist, electrician \\
\bottomrule
\end{tabular}
% }
\caption{\textbf{Full list of occupations used for evaluating biases on FairFace and SocialCounterFactuals datasets.}}
\label{tab:occupations}
\end{table*}

\begin{table*}
\centering
    \begin{tabular}{c|c}
    \toprule
        \textbf{Prompt} & \textbf{Example} \\
        \toprule
        A <occupation> & A biologist\\
        A photo of <occupation> & A photo of biologist\\
        A picture of <occupation> & A picture of biologist\\
        An image of <occupation> & An image of biologist\\
    \bottomrule
    \end{tabular}
    \caption{\textbf{Prompts used for measuring biases on FairFace and SocialCounterFactuals datasets.}}
\end{table*}

\end{document}